\documentclass[10pt,twocolumn,letterpaper]{article}

\usepackage{wacv}
\usepackage{times}
\usepackage{epsfig}
\usepackage{graphicx}
\usepackage{amsmath}
\usepackage{amssymb}
\usepackage{placeins}

\usepackage{algorithm}
\usepackage[noend]{algpseudocode}
\usepackage{subfig}
\usepackage{balance,authblk}
\usepackage{txfonts}
\usepackage{mathdots}
\usepackage[classicReIm]{kpfonts}
\usepackage{nth}
\usepackage{array}
\usepackage{booktabs}
\usepackage{makecell}

\wacvfinalcopy 

\def\wacvPaperID{1} 

\ifwacvfinal\fi
\begin{document}

\title{Structured GANs}


\author[1]{Irad Peleg}
\author[1,2]{Lior Wolf}

\affil[1]{Tel Aviv University}
\affil[2]{Facebook AI Research} 

\maketitle
\ifwacvfinal\thispagestyle{empty}\fi

\begin{abstract}
We present Generative Adversarial Networks (GANs), in which the symmetric property of the generated images is controlled. This is obtained through the generator network's architecture, while the training procedure and the loss remain the same. The symmetric GANs are applied to face image synthesis in order to generate novel faces with a varying amount of symmetry. We also present an unsupervised face rotation capability, which is based on the novel notion of one-shot fine tuning.
\end{abstract}

\section{Introduction}

Symmetry is a prominent property that has gained attention in face recognition~\cite{pentland1994view,adini1997face,brunelli1993face} and other applications~\cite{DBLP:conf/icip/LiangCLB16,cohen2016group}.   
Considering symmetry in the context of image generation with GANs, the core research questions that we ask in this work are: (1) how to control the symmetry of generated images and (2) how to rotate an object, when the training set did not contain the relevant supervision.

Question number one is answered by proposing two alternative  architectures for generative networks. In the first architecture, the first few elements of the input vector serve as the antisymmetric component, the others serve as the symmetric component. In the second architecture, the generated image is symmetric, if the input vector has a palindrome structure, i.e., remains unchanged when flipping the order of elements. Both architectures are shown to work much better than an approach in which the loss is used in order to control the symmetry property. Fig.~\ref{fig:meth1n2Examples} illustrates how $z$ is converted to $G(z)$ in the symmetric GANs.

The second question is answered by a process in which the generator network is adapted in order to generate a specific face. This powerful technique for preserving a given identity in the generated images is a general one and can be applied to many other GAN-based methods.

As far as we know, we are the first to manipulate the structure of the generator in order to enforce properties on the generated image. Due to space constraints, we have placed in the appendices, a different structure manipulation that creates tilable textured patches. Therefore, the core idea of our work also applies to completely different tasks.


\begin{figure*}
\begin{tabular}{cc}
\includegraphics[width = .47\linewidth]{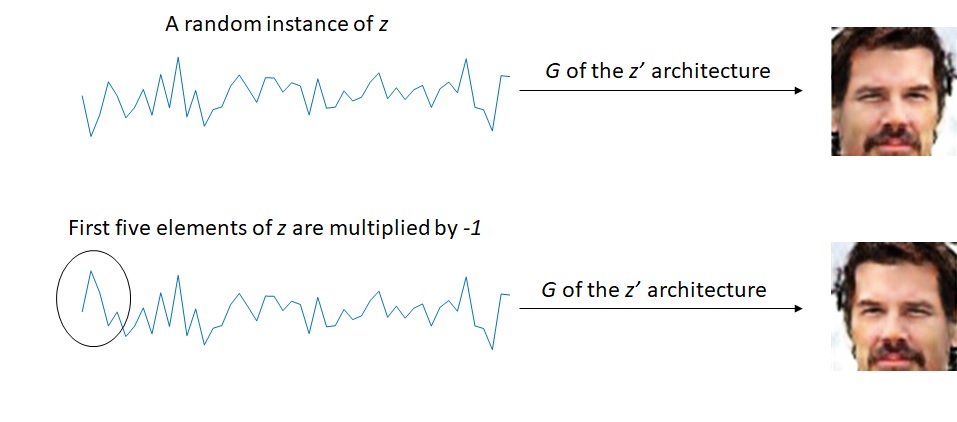} &
\includegraphics[width = .47\linewidth]{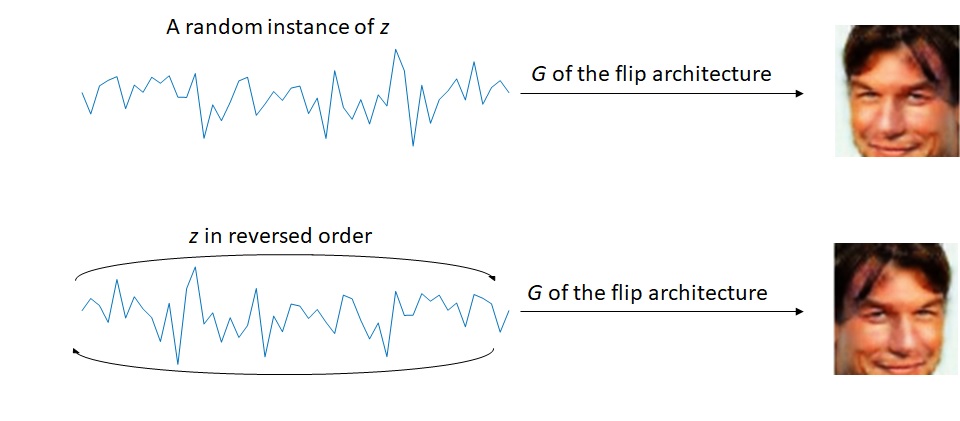}\\
(a)& (b)\\
\end{tabular}
\caption{(a) An illustration of the {\em z' architecture}. The top plot depicts a random instance of $z$, which passes through a generator network to create a face image. The bottom plot shows the vector $z$ after the first five elements multiplied by $-1$. Passing it through the generator of the $z'$ architecture, results with a mirrored version. (b) An illustration of the symmetric GAN {\em flip architecture}. The top plot depicts a random instance of $z$, which passes through a generator network to create a face image. The bottom plot shows $z$ which is left-right flipped. The application of the generator of the flip architecture on this vector results with a mirrored face image.}
\label{fig:illustrate}
\end{figure*}

\begin{figure*}
\includegraphics[width = \linewidth]{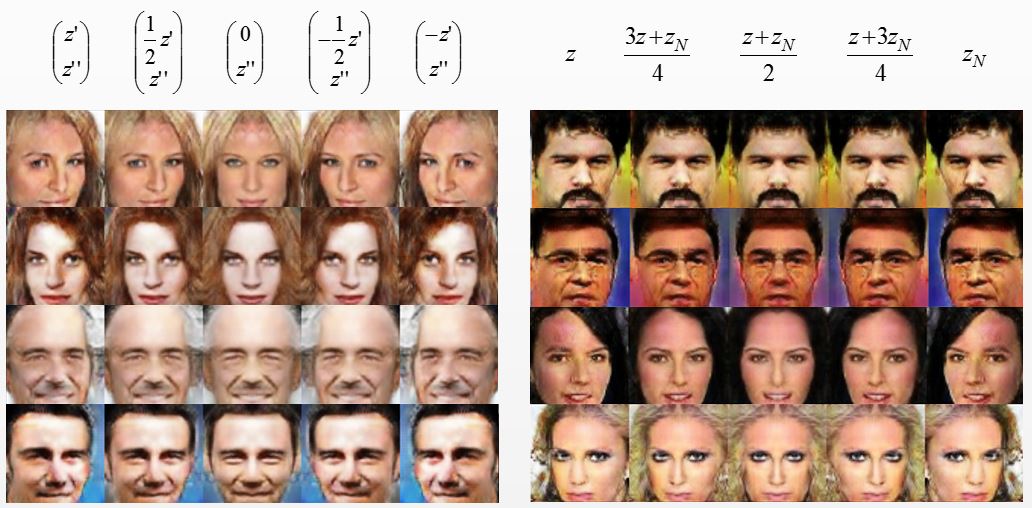}
\caption{An illustration of the relationship between $z$ and $G(z)$ for the two proposed methods. (Left) In the first method, called the $z'$ architecture, part of the GAN's input encodes symmetry. When this part $z'$ is zero, the face is symmetric, and when $z'$ is negated, the image is mirrored along the y axis. (Right) In the flip architecture, the image is symmetric, when the flipped version of $z$, which we denote as $z_N = \textbf{flip}(z)$ satisfies $z_N = z$.}
\label{fig:meth1n2Examples}
\end{figure*}

\subsection{Related Work}

The task of generating realistic looking images has challenged computer vision for a long time. Recently, a major leap has been made with the development of the Generative Adversarial Network (GAN)~\cite{goodfellow2014generative}. These architectures employ two  networks $G$ and $D$, which provide training signals to each other. The network $D$ tries to distinguish between the ``fake'' images generated by $G$ and the real images provided as a training set. Network $G$ tries to fool $D$ and creates images that look as realistic as possible. 

The specific architecture that we employ is based in part on the DC-GAN~\cite{dcgan} method. This architecture uses deconvolution and batch normalization in order to create attractive outputs. Specifically, the input vector in DC-GAN is a $100D$ vector, whose elements are sampled i.i.d from a uniform distribution. In our case, we encode symmetry into this vector, and through the use of specific architectures, we enforce the generated output $G(z)$ to display the required level of symmetry.

Through our focus on symmetry, our networks are able to extract the notion of yaw. GAN-based ways to extract semantic regularities in an unsupervised manner, include InfoGAN~\cite{8136173} and Fader networks~\cite{fader}. 
 
\section{Symmetric GANs}

We present two architecture-based methods, which differ in the manipulation that the input $z$ undergoes in order to create a mirrored version, see Fig.~\ref{fig:illustrate}. We also present a loss-based method to serve as a baseline. The difficulty in training this loss-based method successfully, emphasizes the effectiveness of the architecture-based methods.

\subsection{The Symmetric Architectures}

Our GANs are symmetric end-to-end, including both the generator $G$ and the discriminator $D$.

\begin{figure}
\includegraphics[width = \linewidth,trim={4cm 10cm 4cm 5cm},clip]{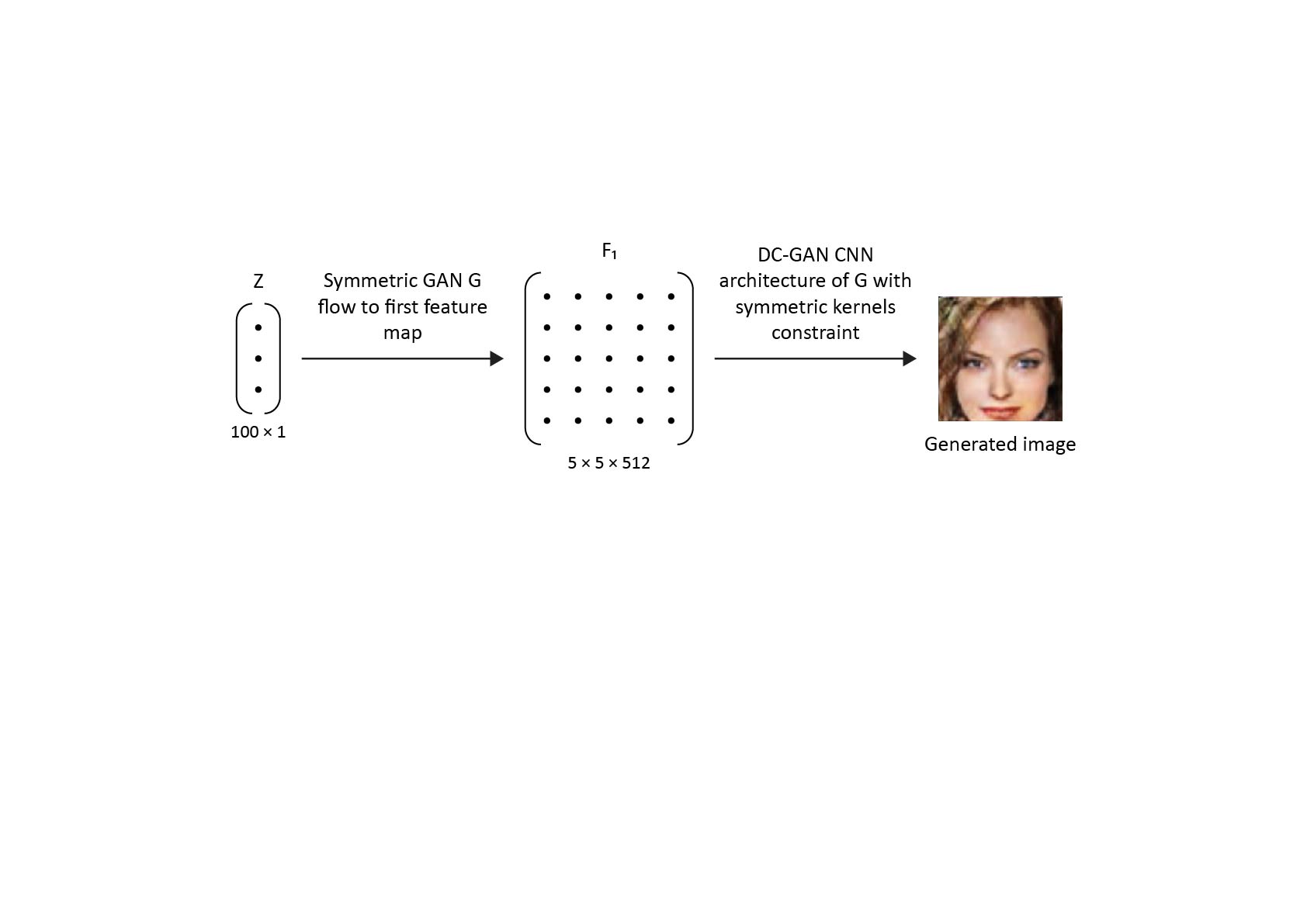}
\caption{Flow of symmetric generator.}
\label{fig:sym_G_flow}
\end{figure}

Both $G$ and $D$ contain convolutional layers, as well as fully connected ones. In order to maintain symmetry, both these layer types are augmented. The fully connected parts are handled differently in $D$ and in $G$. We also present two architectures for $G$, which differ exactly in the way in which the fully connected layers are  constructed. The convolutional layers are treated exactly the same in all variants of $G$ and in $D$ and in all of these cases the same symmetric kernels are introduced.
The flow of the symmetric generator in presented in Fig.~\ref{fig:sym_G_flow}.

The two alternative architectures differ in the very first layer. The first architecture creates symmetric images for inputs $z = {z' \brack z''}$ in which the first part $z'$ is zero. The second architecture produces symmetric outputs for inputs $z$ which are themselves symmetric, i.e., $z = \textbf{flip}(z)$, where the $\textbf{flip}$ operator switches the first element with the last one, the second element with the one before the last and so on. 

\subsubsection{The Generator of the z' Architecture}

The first generated architecture splits the input vector $z$ into two parts. The first, $z'$ is the anti-symmetric part, while the second $z''$ is the symmetric part. For input vectors that have $z'=0$, the output is completely symmetric. 

The architecture that ensures the symmetry and anti-symmetric property enforces this structure on the first feature map, and maintains it thereafter by employing symmetric convolutional kernels.

\begin{figure}
\includegraphics[width = \linewidth,trim={2cm 3cm 2cm 2cm},clip]{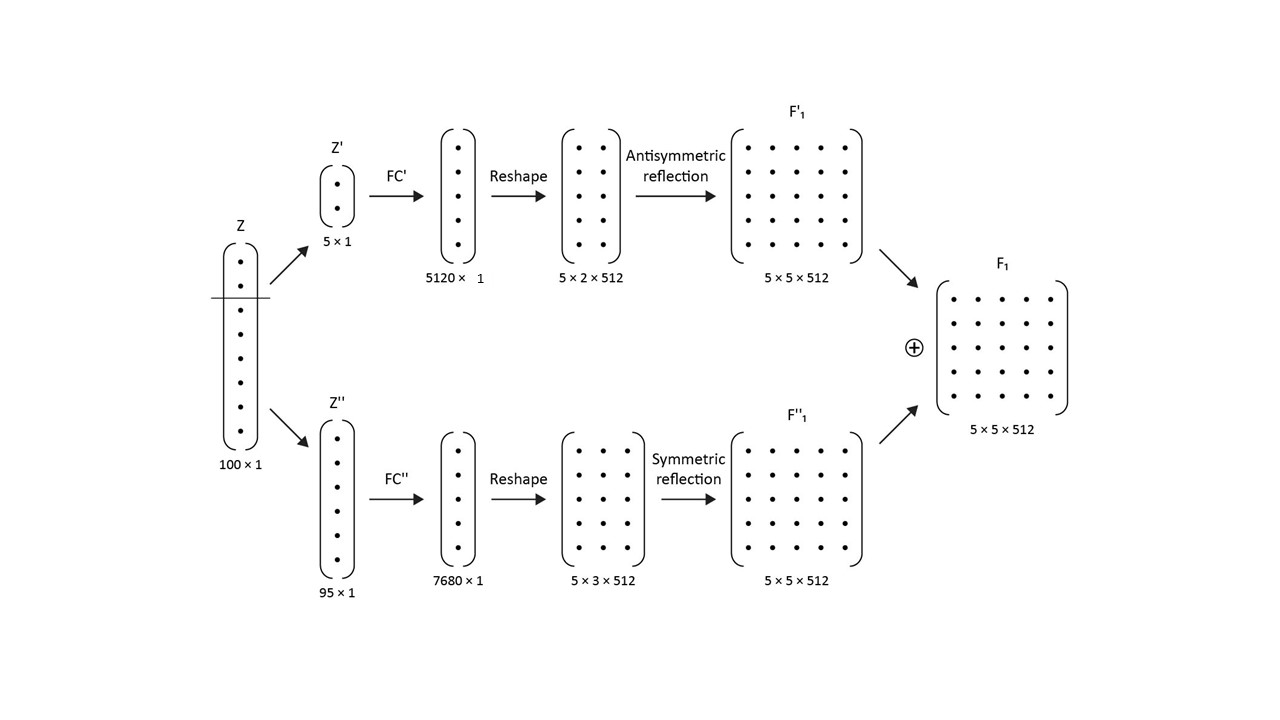}
\caption{Symmetric GAN G flow to first feature map decomposed to symmetric and antisymmetric components}
\label{fig:symmetric_G_method_zprime}
\end{figure}

\begin{figure}
\includegraphics[width = \linewidth,trim={3.5cm 11cm 4.2cm 3cm},clip]{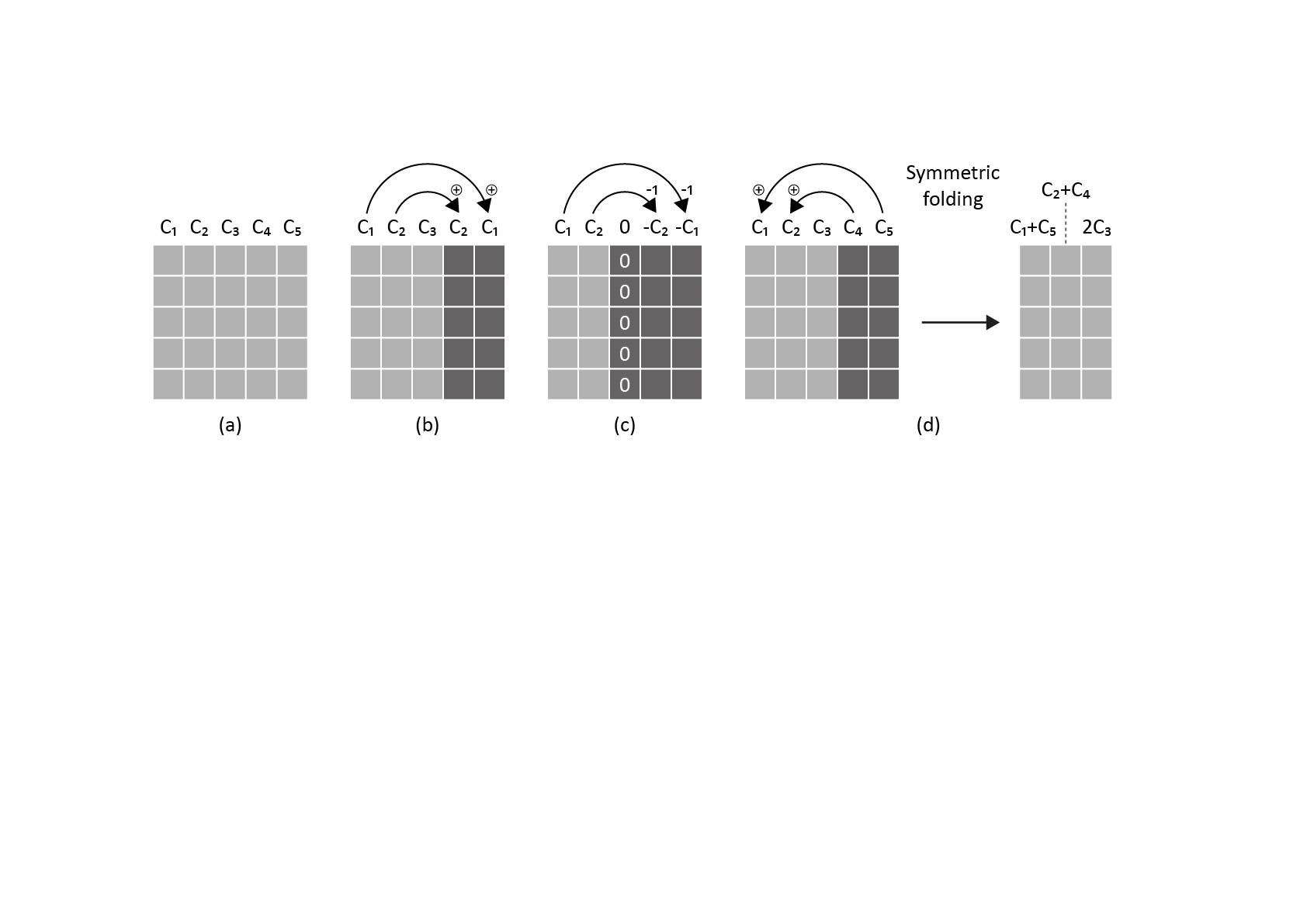}
\caption{A comparison of four kernel types. (a) The standard kernel or feature map. (b) The symmetric kernel and feature map. (c) The anti-symmetric feature map. (d) Symmetric folding of a feature map.}
\label{fig:symmetric_operators}
\end{figure}

The generation of the first feature map is depicted in Fig.~\ref{fig:symmetric_G_method_zprime}. The random input vector $z$, which contains 100 i.i.d uniformly $[-1,1]$ distributed elements, is split into two parts. The first part $z'\in \mathbb{R}^5$ is mapped through a fully connected layer (affine transformation) to a vector of size 5120. This vector is then reshaped into 512 kernels of size $5 \times 2$, which are transformed to antisymmetric kernels of size $5 \times 5$ by taking the last column to be the negative of the first column and the column before, to be the negative of the second column, see Fig.~\ref{fig:symmetric_operators}(c).

The remaining 95 elements of $z$, denoted by $z''$ are mapped to a vector of size $7,680$, which is then reshaped to 512 kernels of size $5 \times 3$. By performing the symmetric reflection depicted in Fig.~\ref{fig:symmetric_operators}(b), i.e., copying the first column to the fifth column and the second to the fourth column, $5 \times 5$ kernels are obtained.

The rest of the network follows the DC-GAN architecture, except that the $5\times 5$ kernels of $25$ parameters each, are replaced with symmetric kernels that contain only 15 different parameters.

Fig.~\ref{fig:symmetric_operators}(c) shows an antisymmetric reflection of the symmetry break part on the first feature map of G.

\subsubsection{The Generator of the $\textbf{flip}$ Architecture}

In the alternative symmetric architecture, the generator produces symmetric images, when for a given input $z$, it happens that $z = \mathbf{flip}(z)$. Here, too, the symmetric kernels throughout the layers make sure that a symmetric feature map from one layer, leads to a symmetric feature map in the next.

\begin{figure}
\includegraphics[width = \linewidth,trim={2cm 3cm 2cm 2cm},clip]{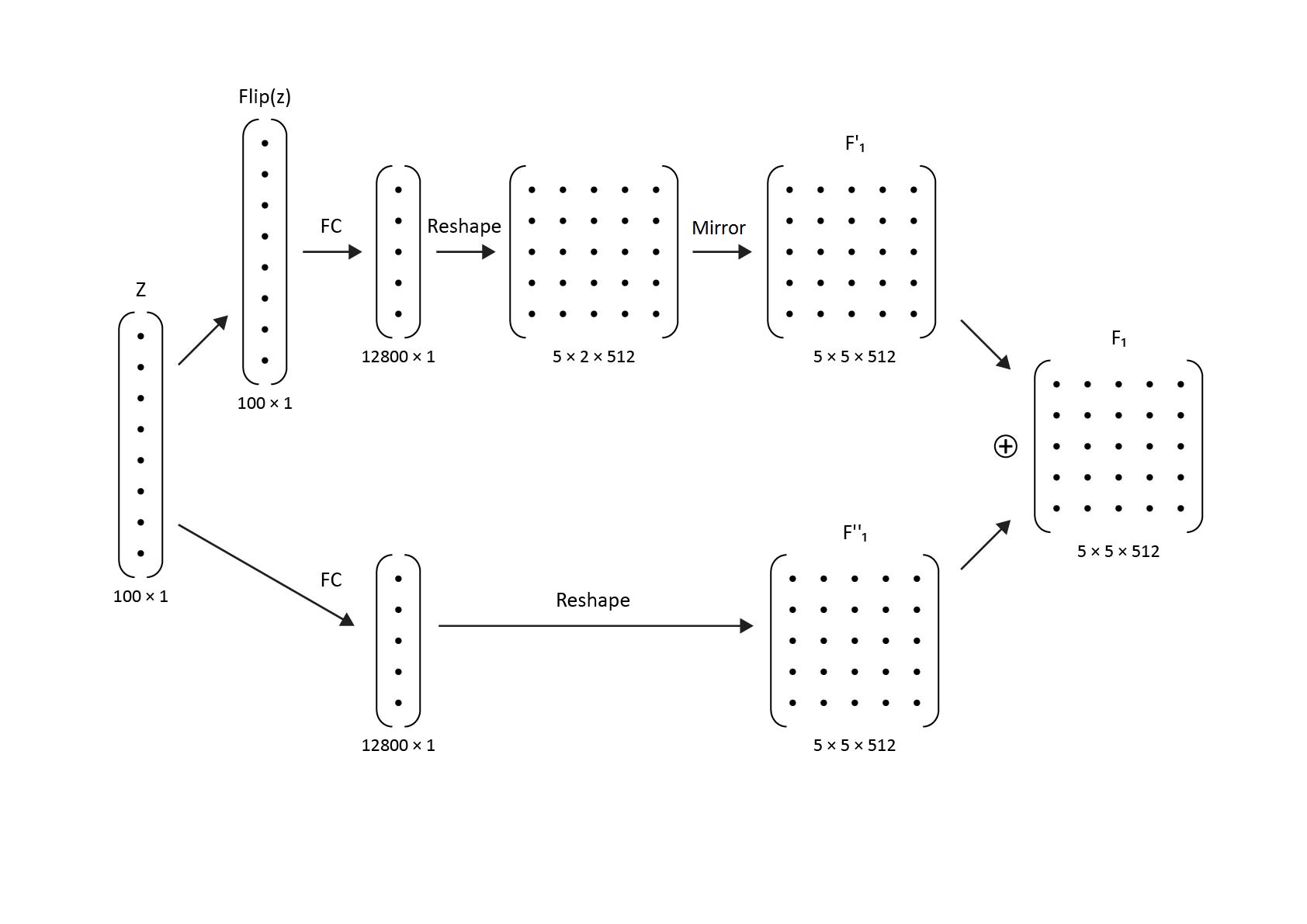}
\caption{symmetric GAN G flow to first feature map as symmetric mapping from a vector to an image.}
\label{fig:symmetric_G_method_flip}
\end{figure}

The architecture of the first layer is depicted in Fig.~\ref{fig:symmetric_G_method_flip}. The same matrix is applied as weights of a fully connected layer, to both $z$ and $\mathbf{flip}(z)$, to obtain vectors of length 12,800. These are reshaped to 512 feature maps of size $5 \times 5$. The feature map that results from the flipped vector is being flipped left and right (the first columns is replaced with the fifth and second with the fourth). The $512$ matching pairs, one from each branch of the network, are then summed. 

Given that $z = \mathbf{flip}(z)$, the two branches produce identical feature maps before the left-right flipping and after flipping, thus symmetry is obtained.

\subsubsection{The Symmetric Discriminator Architecture}
\label{sec:disc}

\begin{figure}   
\centering
\includegraphics[width = \linewidth]{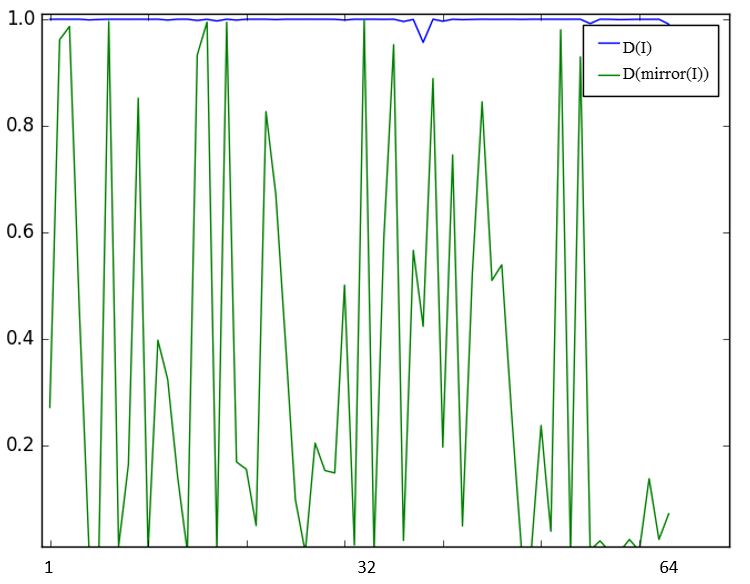}
\caption{The output probability of the ``real'' label of the classifier $D$ for the training samples and their mirrored image, when the classifier does not use the symmetric invariant architecture of Sec.~\ref{sec:disc}. There is a clear overfitting on the training data. However, there is also a great uncertainty about the flipped version.}
\label{fig:Dx_Dxflip}  
\end{figure}

A desired property in our framework is that the discriminator $D$ would return the same probability of ``real'' for an image and its mirrored version. This is not the case in conventional architectures, and, for example, during training the conventional discriminator would overfit on the training sample, while fails on its mirror image, see Fig.~\ref{fig:Dx_Dxflip}.

\begin{figure}
\includegraphics[width = \linewidth,trim={2cm 10cm 2cm 5cm},clip]{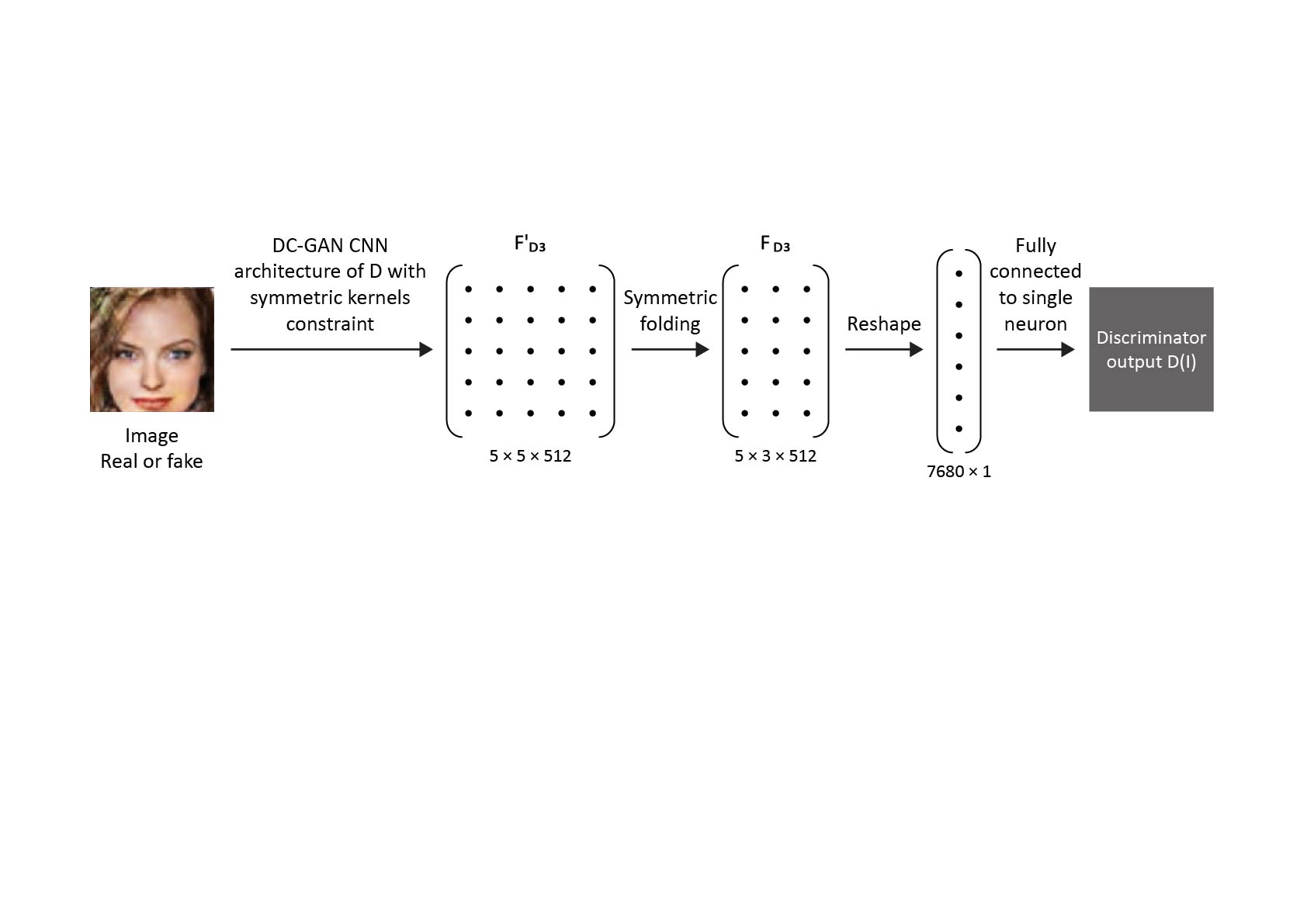}
\caption{Flow of symmetric discriminator.}
\label{fig:sym_D_flow}
\end{figure}

We, therefore, enforce symmetry by using a specific architecture. First, we replace the discriminator of the DCGAN with one that has symmetric left-right kernels, similar to what we have used in the generators. For the last layer, we then obtain a feature map of size $5\times 5\times 512$. This feature map undergoes a symmetric folding, as depicted in Fig.~\ref{fig:symmetric_operators}(d). Namely, the first (second) column is replaced by the sum of this column and the fifth (fourth) column. In addition, the third column is doubled. The result can be seen as a vector the size of $7680$. Through a fully connected layer, a single output is then obtained. The succession of symmetric kernels and the symmetric folding, ensure that mirror images obtain the same score by $D$.
The flow of the symmetric discriminator is shown in Fig.~\ref{fig:sym_D_flow}.

\subsection{Alternative Loss Based Method}

In addition to the architecture based approaches, we evaluated baseline methods for which the symmetry is enforced by adding a loss term. Multiple experiments were conducted, each with varying emphasis (weight) of the added loss term. The overall conclusion is that such training is highly unstable and that the model tends to collapse and either ignore the term (when the weight is low), or result in symmetric images for every input (when the weight is high).

As above, the vector $z$ encodes symmetry either by having the first five elements $z'$ as zero, or by having the vector invariant to element flipping.  

The loss term we add is based on pairs of inputs $z$. For the $z'$ based symmetry encoding, we create pairs $(z,z_N)$, where $z_N$ is identical to $z$, except that the first elements $z'$ are replaced by $-z'$. For the flipping based symmetry encoding, we take $z_N = \textbf{flip}(z)$. The loss term used aggregates over all such pairs $\alpha\|G(z)-\textbf{mirror}(G(z_N))\|^2$, for a weight $\alpha$ and where $\textbf{mirror}$ flips the order of the columns of the image. 

\section{Processing an Existing Image}
\label{sec:GI}

\begin{figure}
\includegraphics[width = \linewidth]{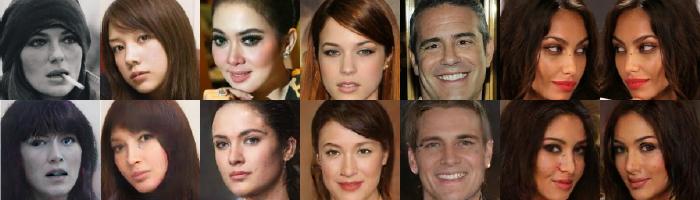}
\caption{Optimizing vector $z$ to fit the top input image in BEGAN~\cite{began} (used with permission). The generated images look realistic and somewhat similar. However, identity is not preserved.}
\label{fig:beganImage}
\end{figure}

In order to manipulate an existing image $I$, one needs to first recover the ``underlying'' $z$ vector by employing a reconstruction loss. 
Employing this vector in order to recover $I$ and rotated versions of it, suffers from noticeable reconstruction errors. Most notably, the reconstructed face $G(z)$ does not maintain the identity of the image $I$, despite their similarity. The same phenomenon is also apparent when observing the approximation results of the most recent generation schemes, such as BEGAN~\cite{began} (see Fig.~\ref{fig:beganImage}).

We therefore, propose to fine-tune $G$ using the same loss, while focusing on the recovered $z$. By doing so, we obtain an image specific network $G_I$ that is able to generate the input image $I$, but is no longer as general as $G$. 

We assume a symmetric generator that employs the $z'$ method. Having $G_I$ and $z$, we are then able to alter the amount of symmetry by modifying $z$. First, the mirror image is generated by employing $z_N$ (as above). The spectrum in between the image and the mirror image, which provides a virtual yaw effect, is then spanned by interpolating between $z$ and $z_N$. The results are shown in Fig~\ref{fig:Gtransfer}(d). As shown in the experiment, the results are superior to those obtained using the unmodified generator $G$.


\begin{algorithm}[t]
\caption{Recovering the underlying vector $z$ for an input image $I$ and modifying the generator $G$ in order to obtain a version $G_I$ such that $I = G_I(z)$ }\label{algo:giz}
\begin{algorithmic}[1]
\Procedure{$[z,G_I] = $ FitAndTune$(G,I,\alpha,\beta)$}{}\\
I. Solve the following optimization problem:\\
~~~$z \gets \textit{argmin}_z \|G(z)-I\| + \alpha \|z\|^2 + \beta(|z|-1)_{+}$\\
II. Alternate between the following:\\
\begin{enumerate}
\item Symmetric GAN training for $G_I$ and $D$
\item $z,G_I \gets \textit{argmin}_{z,G} \|G(z)-I\| + \alpha \|z\|^2 + \beta(|z|-1)_{+}$, where $G,z$ are initialized as the current $G_I,z$
\end{enumerate}
\EndProcedure
\end{algorithmic}
\end{algorithm}

The process of recovering $z$ of $G_I$ is illustrated in Algo.~\ref{algo:giz}. First, we iteratively optimize $z$ to minimize the following term: $\textit{argmin}_z \|G(z)-I\| + \alpha \|z\|^2 + \beta(|z|-1)_{+}$. We found it necessary to employ weight decay on $z$ and also to use a hinge-loss to encourage the values to remain in the range $[-1,1]$.

In the second phase, we allow $G$ to be optimized as well, creating a version $G_I$ that is tailored to the specific sample $I$. In order to prevent $G_I$ from becoming degenerate and too specific to the problem of reconstructing $I$, we alternated between iterations that are performed on the training data that was used for training $G$ and between optimizing $z$ and $G_I$ to minimize the reconstruction risk.

\section{Experiments}

We present empirical evaluation results for both types of structured GANs studied: $z'$ and flip.

\subsection{Applying Symmetric GANs to Face Images}
For evaluating the symmetric GAN methods, we have compared the following methods: 
\begin{enumerate}
\item A simple DC-GAN with no symmetric properties.
\item DC-GAN with soft symmetric loss term: $\alpha=40$.
\item DC-GAN with strong symmetric loss term: $\alpha=100$.
\item Our Symmetric GAN using the $z'$ Architecture.
\item Our Symmetric GAN using the flip Architecture.
\end{enumerate}

In each method, we generated a series of nine images while attempting to enforce symmetry on it, so that the first image will be a mirror of the last, the second of the one before the last and so on. Since the number of images is odd, the middle image is expected to be symmetric to itself.  The way of obtaining the symmetry is determined by the method and follows the description in Fig.~\ref{fig:meth1n2Examples}. 

\begin{figure}[t]
\includegraphics[width = \linewidth]{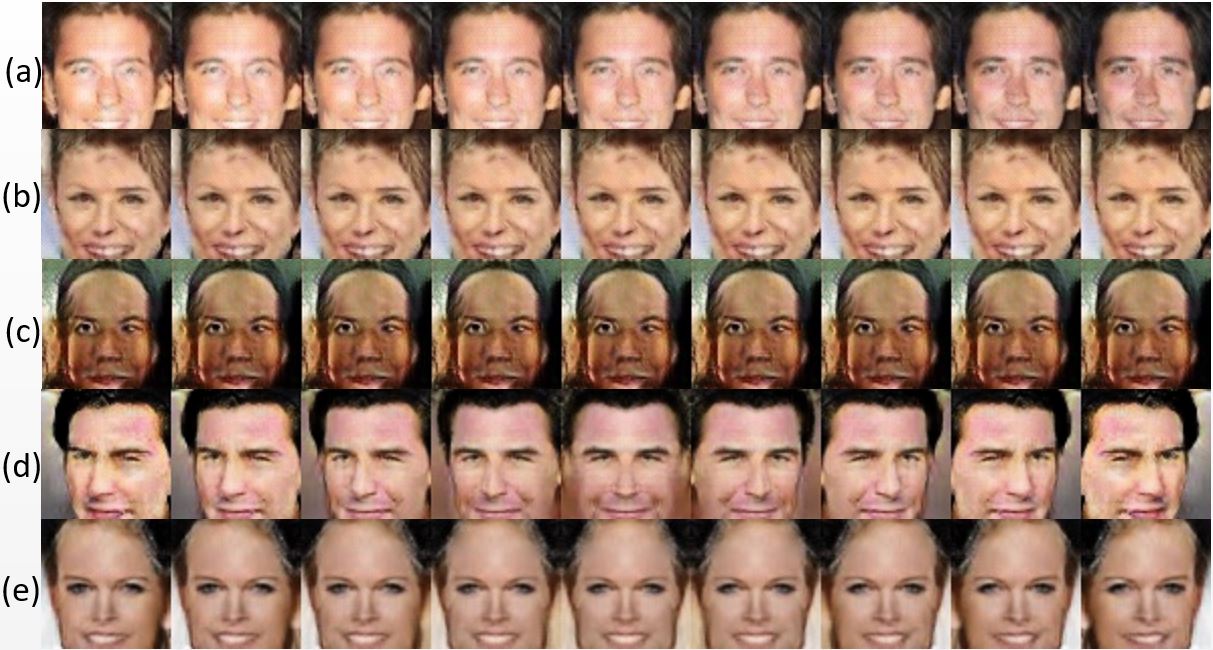}
\caption{Series of images generated by various GAN methods, trying to obtain a mirroring effect between images on the left and right side of each series. (a) DC-GAN. The first 5 elements of the 100D vector $z$ change, but this has little effect on the output image. (b) DC-GAN with a soft symmetric loss term, using the z' encoding of symmetry.(c) DC-GAN with a strong symmetric loss term and using the z' encoding of symmetry. (d) Symmetric-GAN using the $z'$ architecture.(e) Symmetric-GAN using the flip architecture.}
\label{fig:all_methods_strips}
\end{figure}

Sample results are presented in Fig.~\ref{fig:all_methods_strips}. As can be seen, DC-GAN creates high quality images but had no symmetric effect as expected. DC-GAN with soft symmetric loss was not strong enough to enforce symmetry over the generated image. However, on the other hand, it created light deformation to the image caused by unstable training. DC-GAN with a strong symmetric loss was very unstable during training. The generated images were mostly symmetric to themselves and with bad quality. The results of both of our symmetric training techniques were much more convincing and presented the desired effect.

\begin{figure}
\includegraphics[width = \linewidth]{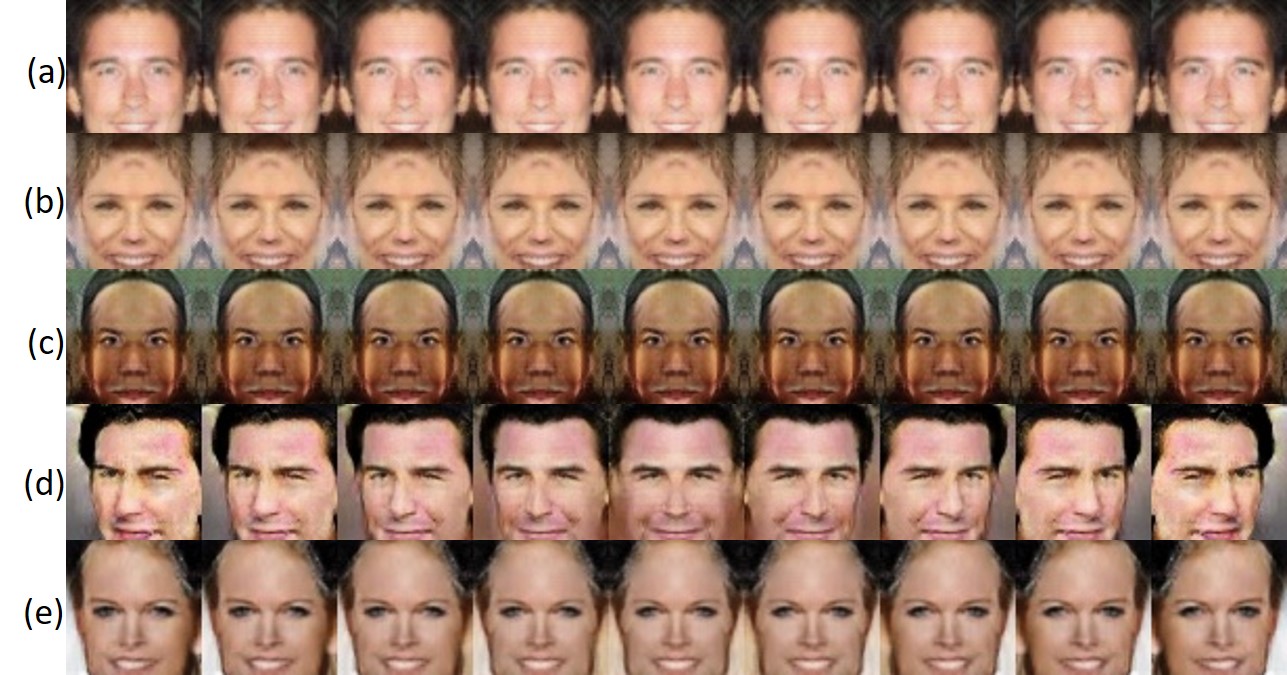}
\caption{Same as Fig~\ref{fig:all_methods_strips} where symmetry is further evaluated by averaging the image $G(z)$ with the mirror image of $G(z_N)$, where $z$ is manipulated to generate the mirror image, i.e., in each location we present $(G(z)+\textbf{mirror}(G(z_N)))/2$. (a) DC-GAN. (b) DC-GAN with a soft symmetric loss term, z' encoding of symmetry.(c) DC-GAN with a strong symmetric loss term, z' encoding. (d) symmetric-GAN using the $z'$ Architecture. (e) Symmetric-GAN using the flip Architecture.}
\label{fig:face_on_face_dcgan}
\end{figure}

We then averaged each generated image with the corresponding image on the other side of the series, after the 2nd image was mirrored. If the mirroring effect is exact, no artifacts are expected. The results can be seen in Fig.~\ref{fig:face_on_face_dcgan}. This visualization clearly demonstrates that both our methods ($z'$ and flip) create mirror images when the input dictates this.

\begin{figure*}
\centering
\includegraphics[width = 0.96\linewidth]{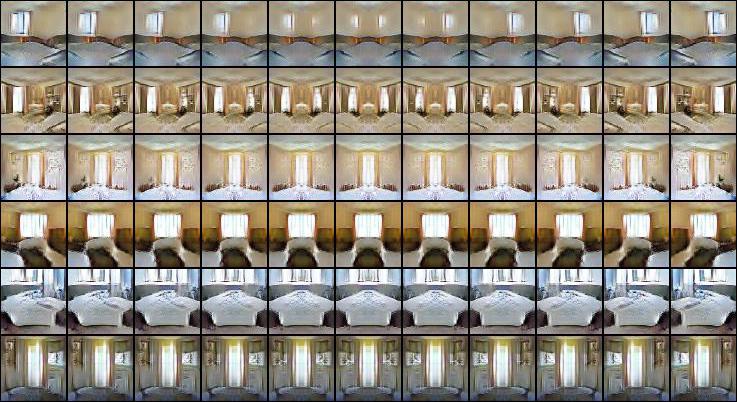}
\caption{Using the $z'$ architecture, $z'$ is multiplied by a scalar, shifted from [-0.5,+0.5] with leaps of 0.1 (left to right). For each row, $z''$ is held constant. Each row is a different experiment, producing 11 images.}
\label{fig:lsun5D}
\end{figure*}

Finally, we measure the MSE between each image and the mirror version of it. The results are shown in Fig.~7 in the appendices. As can be seen, the proposed methods drop to nearly 0 in the middle image, indicating that those images are symmetric. We can see that the MSE of the other methods is relatively constant and does not drop to zero. The loss-based method, with the strong symmetric constraints creates images that are symmetric throughout the range of $z'$ values. An even stronger symmetry loss would lead to an MSE close to zero along the entire curve, with an image that is barely recognizable as a face.

\begin{figure*}[t]
\centering
\scalebox{1.452}{
\subfloat[]{\includegraphics[width=.1823in]{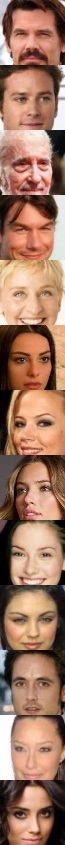}} 
\hspace{0.0005cm}
\subfloat[]{\includegraphics[width=.1823in]{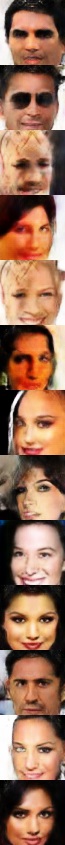}} 
\hspace{0.0005cm}
\subfloat[]{\includegraphics[width=2.735in]{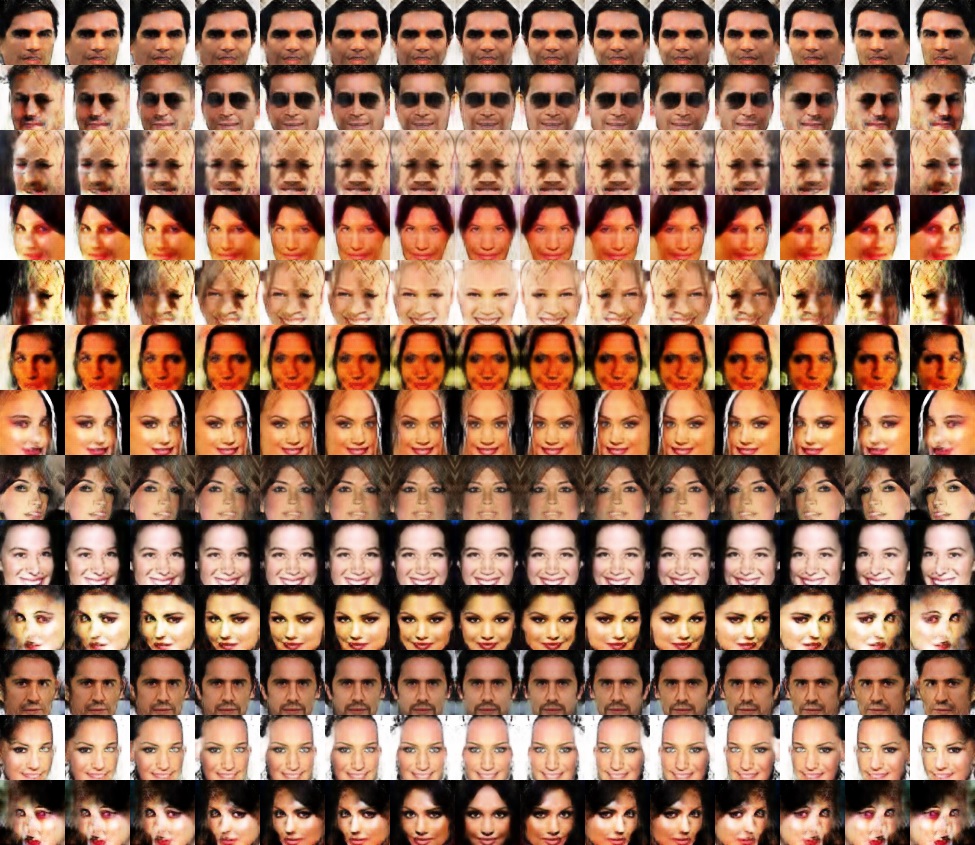}}
}
\caption{The results obtained without $G$ transfer. (a) Sample dataset images $I$. (b) $G(z)$ for the vector $z$ that was optimized to minimize the reconstruction loss. Results are shown for a symmetric generator that is based on the $z'$ architecture. (c) Rotations performed using $G$ and manipulated versions of $z$.\label{fig:Ztransfer}}
\end{figure*}

\begin{figure*}[t]
\centering
\scalebox{1.452}{
\subfloat[]{\includegraphics[width=.1764in]{GT/allorigFaces.jpg}} 
\hspace{0.0002cm}
\subfloat[]{\includegraphics[width=.1764in]{GT/allFaces_init.jpg}} 
\hspace{0.0002cm}
\subfloat[]{\includegraphics[width=.1764in]{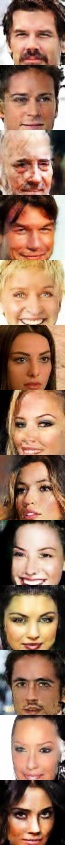}}
\hspace{0.0002cm}
\subfloat[]{\includegraphics[width=2.647in]{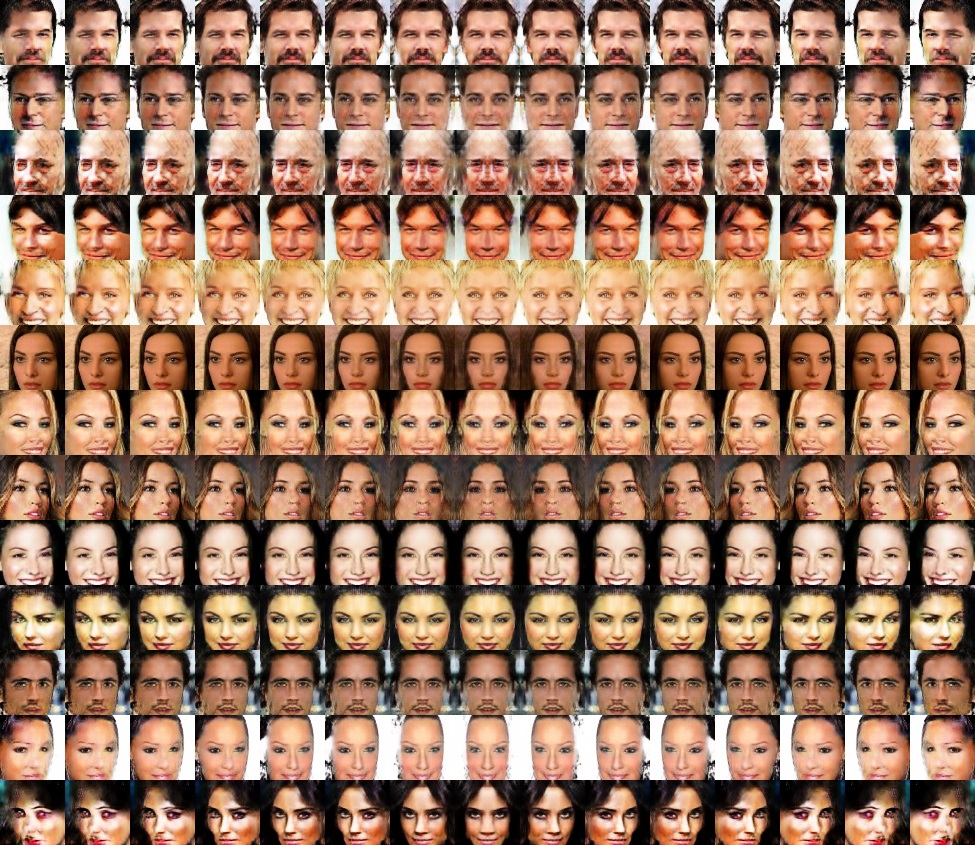}}
}
\caption{(a) Dataset samples $I$. (b) $G(z)$ for the vector $z$ that was optimized to minimize the reconstruction loss. Results are shown for a symmetric generator that is based on the $z'$ architecture. (c) $G_I(z)$, where $G_I$ is finetuned from $G$ in order to further minimize the reconstruction error. (d) Rotations performed using $G_I$ and manipulated versions of $z$.\label{fig:Gtransfer}}
\end{figure*}

\paragraph{Manipulating a Face Image} In order to manipulate a given image $I$, we have recovered the $z$ vector that best matches the image and then manipulated it, as explained in Sec.~\ref{sec:GI}. Fig.~\ref{fig:Ztransfer} depicts the results obtained by recovering this vector and then generating images from manipulated versions. There are noticeable artifacts. These artifacts are largely reduced when performing the per image tuning of $G$ in order to obtain $G_I$, as can be seen in Fig.~\ref{fig:Gtransfer}.



\begin{figure*}
\centering
\includegraphics[width = .7\linewidth]{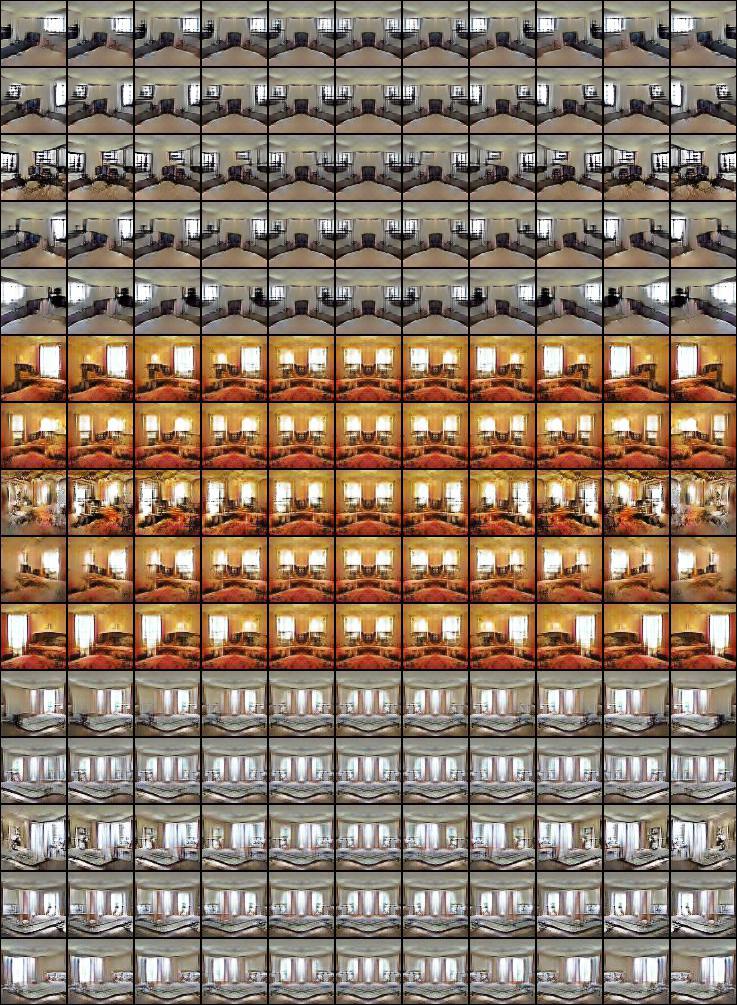}
\caption{Using the $z'$ architecture, all five $z'$  fields, except one, are set to zero. The remaining field is shifted from [-1,+1] with leaps of 0.2 (left to right). In each row, a different field is changed. All 95 $z''$ fields are unchanged during the experiment. Each experiment produce 5x11 images. The following experiment has been repeated three times.}
\label{fig:lsun1D}
\end{figure*}

\subsection{Symmetrical Views of Man-Made Scenes}

To show that our method is general, the network was also trained on the LSUN bedrooms dataset~\cite{yu2015lsun}. Unlike a face, a bedroom is not symmetric. However, since mirror images of rooms belong to the same class,  the method fits this kind of data well. 

In our experiments, we focused on the $z'$ architecture. The first experiment shows how a generated image is affected when setting its $z'$ component closer or further away from zero. The results, depicted in Fig.~\ref{fig:lsun5D}, show that the closer to zero, the more symmetric the generated image is, and that the image associated with $z'$ and that associated with $-z'$ are mirror images. The second experiment is similar, with the single change that we fix $z'=0$ everywhere except for one coordinate at a time that is changed. This way, we can study the effect of each individual dimension on the output. The results are shown in Fig.~\ref{fig:lsun1D}. It is clear that each dimension controls a different mode of variability. However, the dimensions are not independent and the same objects emerge by using different coordinates.

\section{Conclusions}

DC-GANs are being used today for a wide range of  applications, such as domain transfer networks~\cite{taigman2016unsupervised}, photo editing~\cite{brock2016neural}, denoising, data creation and more. We demonstrate how by manipulating the structure of the generator, we can directly control the symmetry of the output. A second, completely different, application to tiling, which is presented in the appendices, shows that a similar structure modifying design provides a solution for a completely different application. 

\section*{Acknowledgements}

This project has received funding from the European Research Council (ERC) under the European Union's Horizon 2020 research and innovation programme (grant ERC CoG 725974).

We are grateful to Barak Itkin for proposing the tiling application.

\bibliographystyle{plain}
\bibliography{gans}

\appendix
\section{Generating Tiles}

As a second application for manipulating the structure of GANs, we present methods for creating tiles that can be arranged repeatedly in 2D in a variety of predetermined patterns. Just like symmetry, tiling enforces a specific structure on the output. For example, in the case of simple tiling, where tiles are being placed in the same orientation on a grid, the top (left) part of the tile should merge smoothly with the bottom (right) part.

\subsection{Previous Work On Texture Synthesis} 
Gatys~\cite{Gatys:2015:TSU:2969239.2969269} demonstrated how to capture texture properties from a given image and generate new images with the same texture properties. The descriptor is based on a pre-trained network, usually VGG~\cite{simonyan2014very}. A GRAM matrix is extracted from feature maps of certain layers. The objective compares the descriptors of the target image to those of the source image.  
~\cite{gatys2015neural,ulyanov17improved} perform style transfer by combining a content loss from a feature map of a deep layer of VGG. Later on, works such as ~\cite{johnson2016perceptual,ulyanov2016texture,li2016precomputed,jetchev2016texture,bergmann2017learning} and others, showed how to train generative networks that are able to simultaneously generate images with texture properties that were already embedded in the train process. Works like ~\cite{li2016precomputed,jetchev2016texture,bergmann2017learning} do so as a GAN implementation.

In contrast to previous work, we focus on tiles and not on the textured image. This allows us to develop GANS that create tiles for complex tiling patterns.

\subsection{An Architecture for Generating Tiles}

The idea of enforcing structure by constructing a suitable architecture, as opposed to modifying just the loss, extends beyond symmetry to the problem of tiling. The input to the tiling problem is an image $I$ of some texture. The goal is to synthesize a patch that:
\begin{enumerate}
\item	Has texture properties that are indistinguishable from those of patches from the source image $I$.
\item	Has a periodic structure such that when the patch is concatenated to itself, there is no texture discontinuity in the boundary.
\end{enumerate}

\begin{figure*}
\centering
\includegraphics[width = 0.77\linewidth,trim={3.5cm 5.2cm 3.5cm 4.9cm},clip]{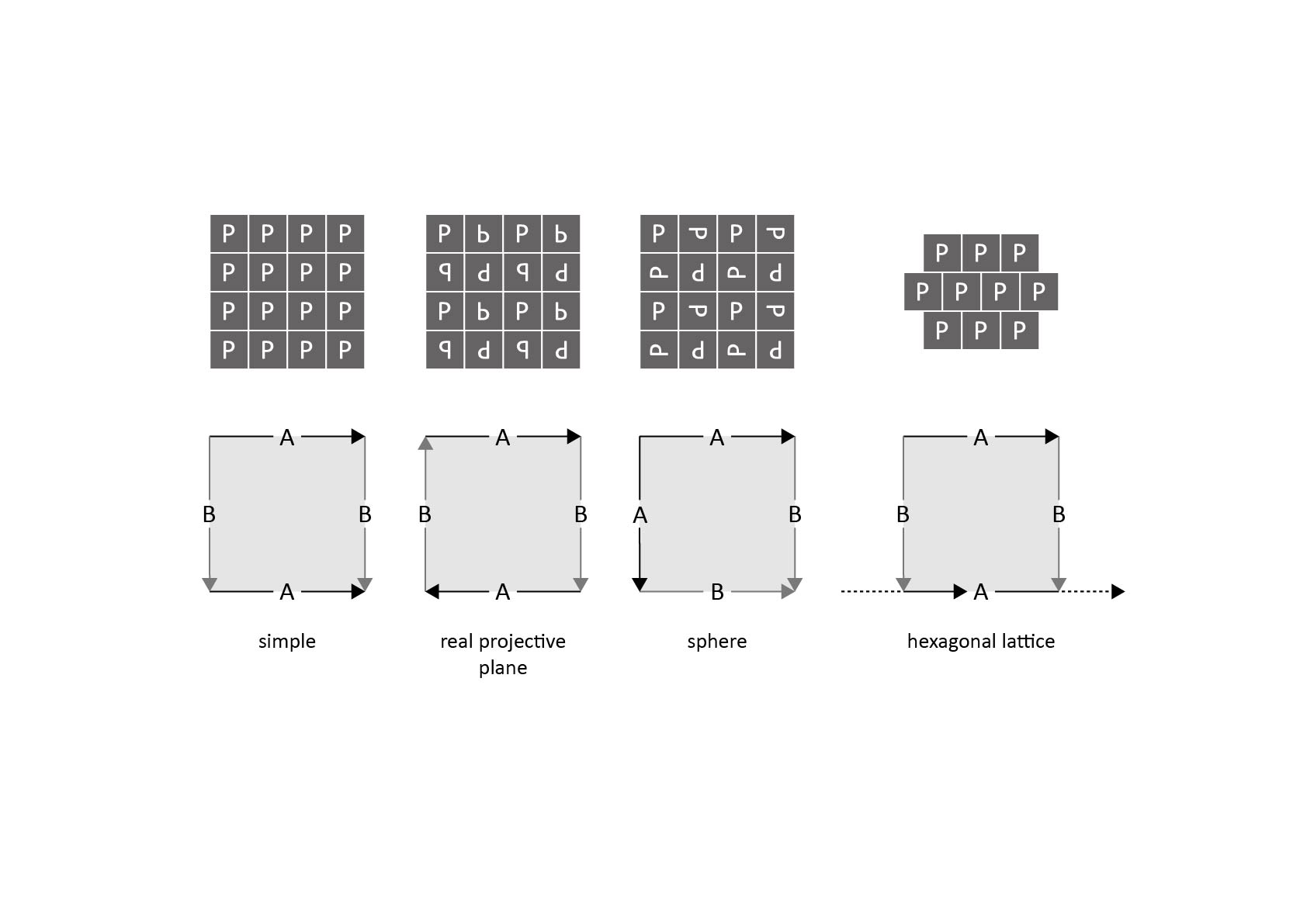}
\caption{The four forms of tiling presented in this paper. From left to right: (simple rectangular lattice, real projective plane topology, spherical topology, hexagonal lattice)}
\label{fig:patterns}
\end{figure*}

The most basic tiling pattern repeats each tile, as is, in multiple columns and rows. However, as Fig.~\ref{fig:patterns} illustrates, there are many alternative patterns in which the patterns might be rotated or placed in more complex patterns.

As in symmetry, we employ a modified version of the generator $G$ of the DC-GAN method~\cite{dcgan} in order to transform a random vector $z$ into a patch image, in this case of size $64\times 64$. Unlike the symmetry encode case, in which the vector $z$ encodes whether the output image is symmetric or not, for tiling, we expect all outputs to maintain the two desired properties and $z$ is completely random.

Since it is the texture properties of the patch that we are concerned with, we encode the patch using the GRAM matrix extracted from the generated image as well as from all layers of $D$, right after the convolution, and before adding the bias, performing batch normalization and applying ReLU. Specifically,
\[GRAM_{ij}^{l} =\left\langle F_{i}^{l} ,F_{j}^{l} \right\rangle, \]
where $F_{i}^{l}$ denotes the $i-th$ feature map of layer $l$. A virtual layer of ones is added in order to capture first order statistics and the size of the GRAM matrix computed for layer $l$ is, therefore, $(k_l+1)^2$, where $k_l$ is the number of filters in this layer.
All GRA matrices are then normalized by the value $k_l^{1.5}$.

\begin{figure*}
\centering
~~~\includegraphics[width = \linewidth,trim={0.5cm 1.5cm 0.1cm 1.5cm},clip]{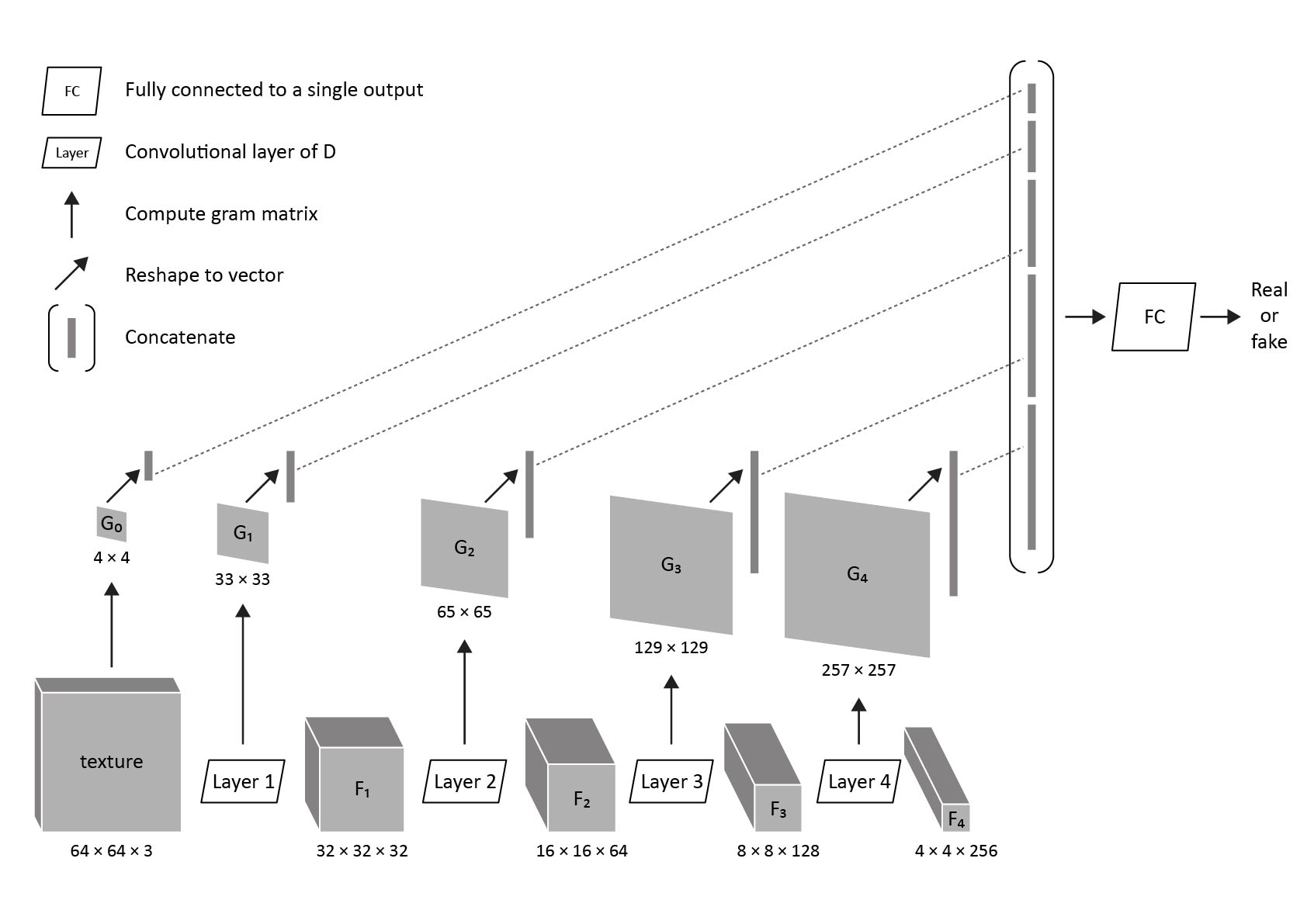}
\caption{The architecture of $D$ for texture synthesis.}
\label{fig:D_texture}
\centering
\subfloat[]{\includegraphics[width = .67\linewidth,trim={4cm 7cm 4cm 6cm},clip]{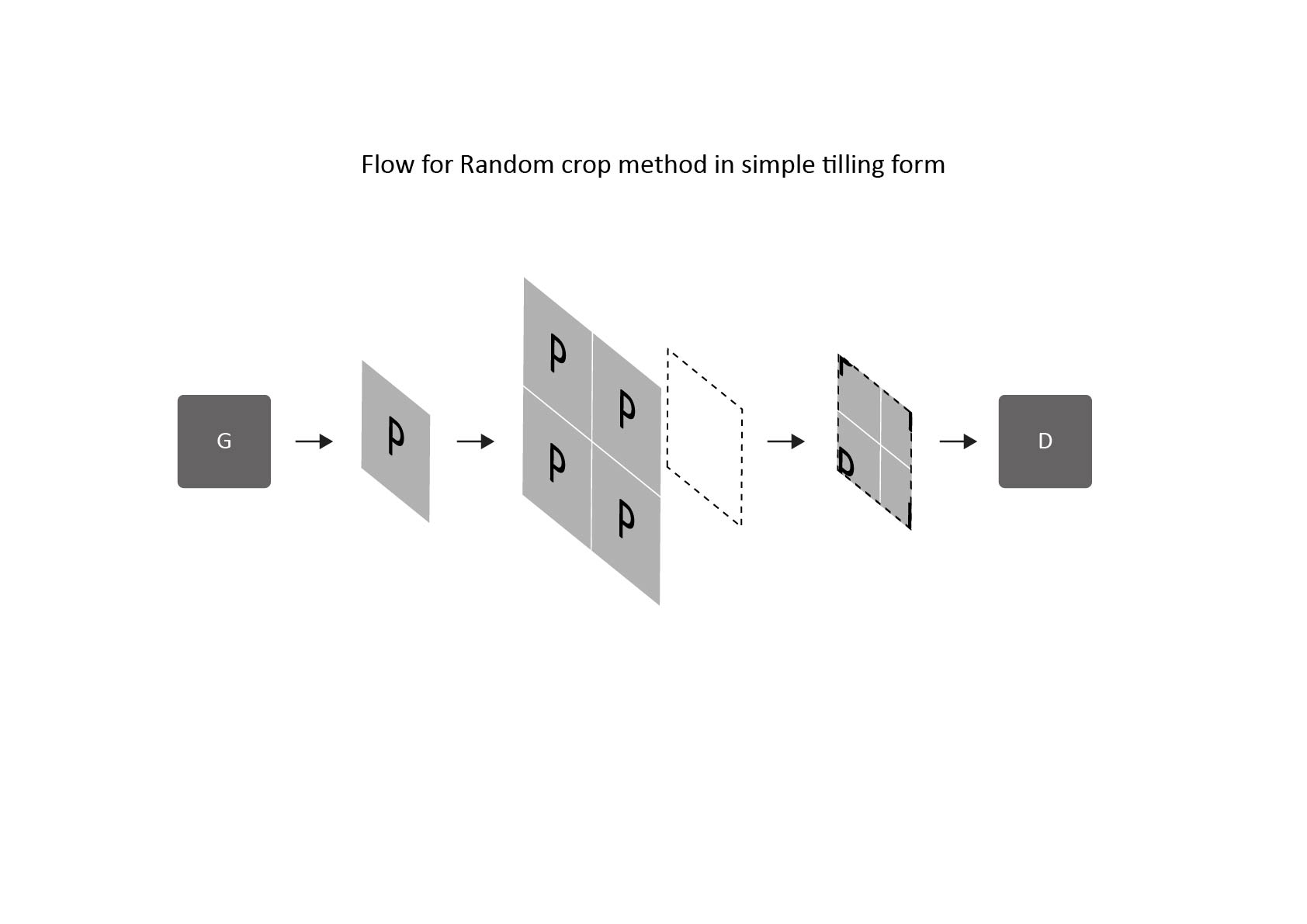}}\\
\centering
\subfloat[]{\includegraphics[width = .67\linewidth,trim={3cm 4cm 3cm 4.2cm},clip]{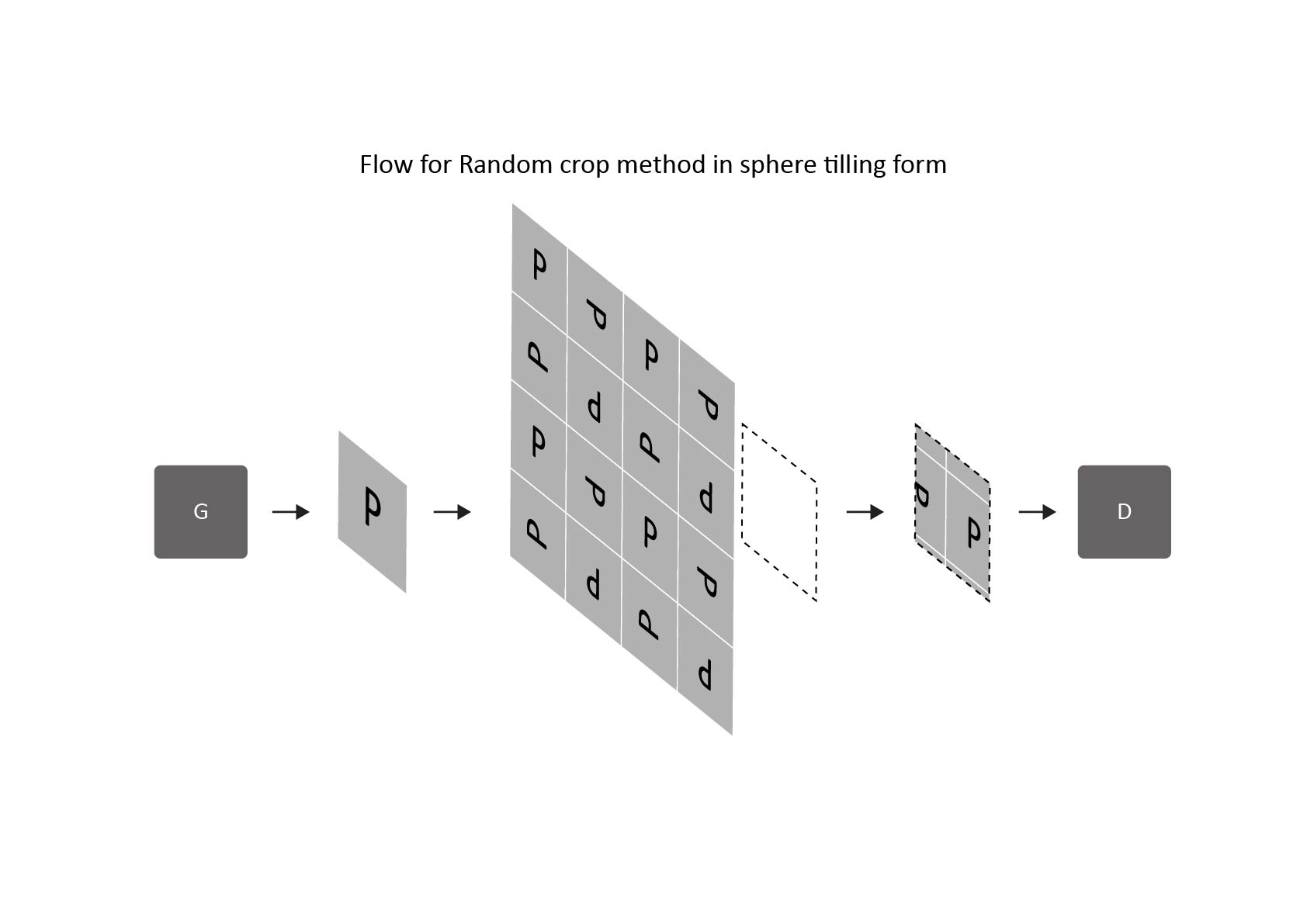}}
\caption{The tile and random crop method as applied to (a) Vanilla tiling on a grid. (b) tiling the spherical topology.}
\label{fig:randCropFlow}
\end{figure*}

All GRAM fields from all the layers of $D$ are concatenated to one descriptor, which is fed to the fully connected part of $D$. At each batch, $64$ crops out of $I$ of size $64\times64 \times 3$ are used as the ``real'' samples and $64$ generated samples of the same size are used as the ``fake'' sample. The architecture of $D$ for capturing textures is depicted in Fig.~\ref{fig:D_texture}. 

We propose two different tiling GAN methods. The first employs cyclic deconvolutions and the second tiles and crops.

\paragraph{Cyclic deconvolution} In order to support horizontal tiling, for example, it is necessary to have the leftmost part of the patch similar to the rightmost part. This is enforced by replacing the deconvolution blocks of $G$ with cyclic deconvolution blocks, in which the convolutions support extend beyond the edges of the feature map and warp back to the other end of the map.  This is done for all layers of $G$. Note, that for complex tiling patterns, the cyclic deconvolution take more complex forms (see below).
\paragraph{Tile and randomly crop} In this method, it is the discriminator that enforces the tiling property. This is done by taking the generated image, tiling it in the plane and cropping a $64 \times 64 \times 3$ patch from the result. This patch is then fed to $D$. If there are tiling artifacts in the crop, the discriminator will then pick up on these. During backpropagation, G is being augmented in a way that reduces the artifacts and learns the tiling pattern implicitly. See Fig.~\ref{fig:randCropFlow}.

\subsection{Tiling experiments}

\begin{figure*}[t]
\includegraphics[width =\linewidth]{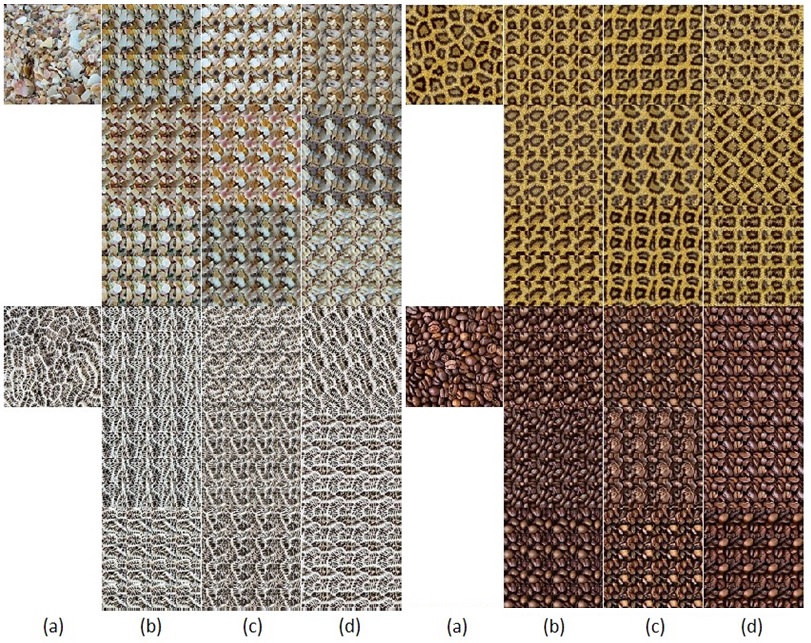}
\caption{A comparison of the tiling methods. (a) real texture. (b) tiling outputs of DC-GAN. (c) the outcome of tiling with the cyclic deconvolution method. (d) the outcome of tiling with the tile and crop method.}
\label{fig:tileBasic}
\end{figure*}

We first present, in Fig.~\ref{fig:tileBasic}, the results obtained for the simple grid tiling. As can be seen, tiling using tiles generated by the baseline DC-GAN leads to noticeable artifacts at the boundaries of the tiles, while either one of the two methods we propose avoids these artifacts.

\begin{figure}[t]
\includegraphics[width = \linewidth]{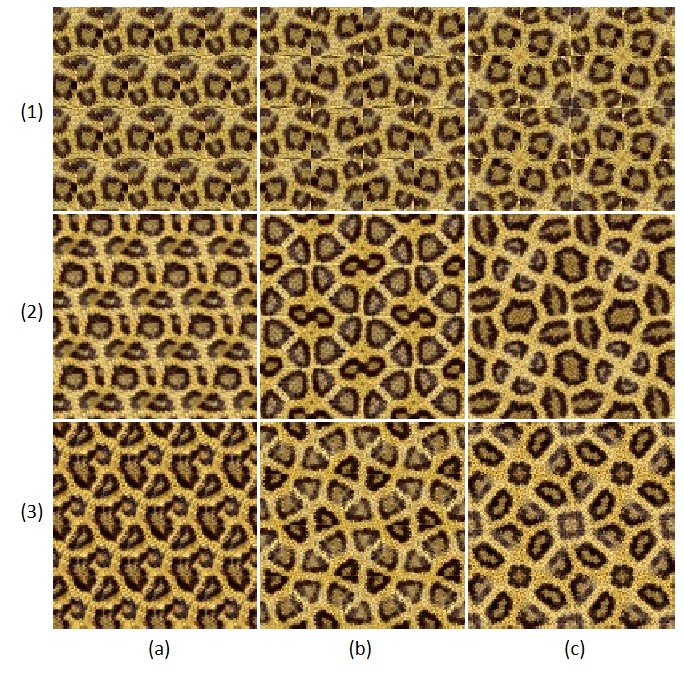}
\caption{Row 1 shows tiling of a real texture. Row 2 shows tiling using the circular convolution method. Row 3 shows tiling of using random crop method. Column a shows tiling in a hexagonal pattern. Column b shows tiling in pattern of real projective plane topology. Column c shows tiling in pattern of spherical topology.}
\label{fig:patternsReal}
\end{figure}

We further experimented with less conventional tiling approaches. The results are shown in Fig.~\ref{fig:patternsReal}. The proposed methods perform well, except that the cyclic convolution method is not appropriate for the spherical topology, since it requires the conversion of a row to a column and vice versa. 

\begin{figure}[t]
\centering
\subfloat[]{\includegraphics[width=1in]{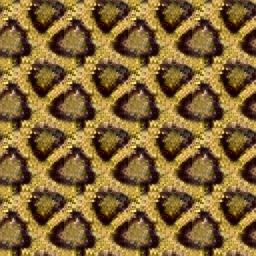}} 
\hspace{0.01cm}
\subfloat[]{\includegraphics[width=1in]{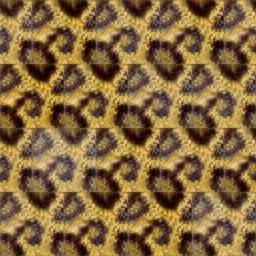}} 
\hspace{0.01cm}
\subfloat[]{\includegraphics[width=1in]{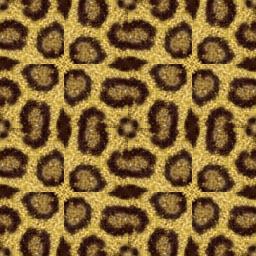}}
\caption{a collection of three repeatable artifacts observed during tiling experiments. (a) a noise texture that appears in some cases of tiling using the random crop method. (b) hexagonal tiling with the tile and crop method results in a constant tile, here each tile has a different $z$ and yet all tiles are the same. (c) a discontinuity phenomenon typical for cyclic deconvolution combined with spherical topology.}
\label{fig:problems}
\end{figure}

A closer look at the various artifacts can be observed in Fig.~\ref{fig:problems}.

\section{MSE Plot for Symmetric GANs}

\begin{figure*}
\centering
\includegraphics[width = 0.8\linewidth]{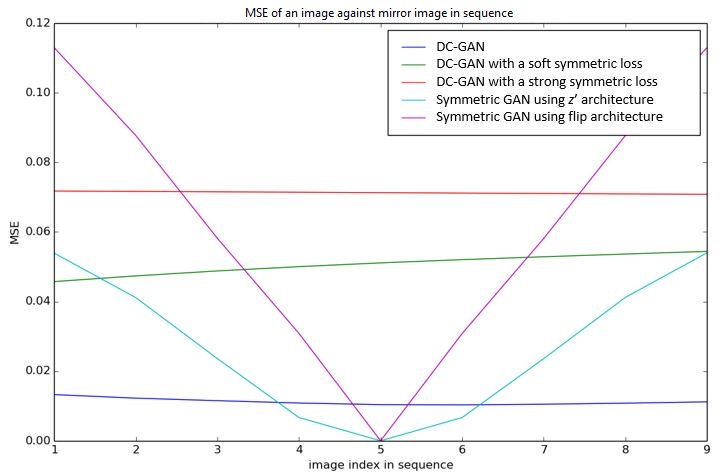}
\caption{ The MSE difference between a generated image $G(z)$ and the mirrored image of $G(z_N)$, where $z_N$ is the vector that is supposed to generate the mirrored image, i.e., for the loss based method and the z' symmetric GAN, the first five coordinates of $z$ are the negative of the first five coordinates of $z'$.}
\label{fig:MSE}
\end{figure*}

We measure the MSE between each image and the mirror version of it. The results are shown in Fig.~\ref{fig:MSE}. As can be seen, the proposed methods drop to nearly 0 in the middle image, indicating that those images are symmetric to themselves. We can see that the MSE of the other methods is relatively constant and does not drop to zero. The loss-based method, with the strong symmetric constraints creates images that are symmetric throughout the range of $z'$ values. An even stronger symmetry loss would lead to an MSE close to zero along the entire curve, with an image that is barely recognizable as a face.

\end{document}